\documentclass[10pt,twocolumn,letterpaper]{article}

\usepackage{isba}
\usepackage{times}
\usepackage{epsfig}
\usepackage{graphicx}
\usepackage{amsmath}
\usepackage{amssymb}
\usepackage{multirow,tabularx}
\usepackage{tabularx}
\usepackage{colortbl}
\usepackage{hhline}
\usepackage{array}
\usepackage{url}
\usepackage{algorithm}
\usepackage{algcompatible}
\usepackage{algpseudocode}
\usepackage{amssymb}
\usepackage{subfig}
\usepackage{stfloats}
\usepackage{wrapfig}
\usepackage{amsthm}
\usepackage{multirow}

\isbafinalcopy 

\ifisbafinal\fi
\begin{document}

\title{FDSNet: Finger dorsal image spoof detection network using light field camera}

\author{Avantika Singh, Gaurav Jaswal, Aditya Nigam\\
Indian Institute of Technology Mandi\\
Mandi, India\\
{\tt\small d16027@students.iitmandi.ac.in, gaurav\_jaswal@projects.iitmandi.ac.in, aditya@iitmandi.ac.in}
}

\maketitle
\thispagestyle{empty}
\begin{figure}[b]
\parbox{\hsize}{\em
2019 IEEE $5^{th}$ International Conference on Identity, Security, and Behavior Analysis (ISBA) \\
978-1-7281-0532-1/19/\$31.00 \ \copyright 2019 IEEE
}\end{figure}

\begin{abstract}
At present spoofing attacks via which biometric system is potentially vulnerable against a fake biometric characteristic, introduces a great challenge to recognition performance. Despite the availability of a broad range of presentation attack detection (PAD) or liveness detection algorithms, fingerprint sensors are vulnerable to spoofing via fake fingers. In such situations, finger dorsal images can be thought of as an alternative which can be captured without much user cooperation and are more appropriate for outdoor security applications. In this paper, we present a first feasibility study of spoofing attack scenarios on finger dorsal authentication system, which include four types of presentation attacks such as printed paper, wrapped printed paper, scan and mobile. This study also presents a CNN based spoofing attack detection method which employ state-of-the-art deep learning techniques along with transfer learning mechanism. We have collected 196 finger dorsal real images from 33 subjects, captured with a Lytro camera and also created a set of 784 finger dorsal spoofing images. Extensive experimental results have been performed that demonstrates the superiority of the proposed approach for various spoofing attacks. 

\end{abstract}

\section{Introduction}
\begin{figure}[!b]
	\begin{center}
		\includegraphics[width=0.98\linewidth]{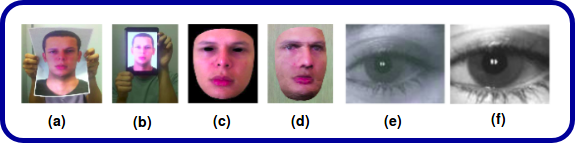}
	\end{center}
\caption{Sample images illustrating various artifacts.  (a) wrapped image, (b) video playback, (c) life-size 3-D mask, (d) paper cut mask (all these images are taken from \cite{facearti}, (e) scanned image (c) captured image (images taken from \cite{7a})}
	\label{fig:1}
\end{figure}

With the increased deployment of biometric based authentication systems in applications like smart-phones, tablets and wearable mobile devices vulnerability of biometric based devices to the presentation attacks is increasing day-by-day \cite{1}. The sole aim of various presentation attack detection algorithms is to thwart the biometric system in order to get an unauthorized access by some attacker \cite{1b1}. When a fabricated material is used to create a spurious biometric modality, it is more commonly known as spoofing attack \cite{1c}. This kind of attack is common in fingerprints or other modalities. Fingerprint spoofs are generated using variety of materials like  Ecoflex, Playdoh, Latex and many more \cite{13s1}. Spoofed materials generated using different materials have different characteristics and it has been shown that methods present in literature are not generalizable to unseen spoof materials on which they are not trained. The existing spoof detection methods can be grouped into two main categories (i) hardware based solutions (ii) software based solutions. The hardware based solutions typically detect physical characteristics like flow of the blood whereas software based solutions typically extract physiological and anatomical features from any biometric trait \cite{3}, thus these software based solutions are much cheaper as they require no-additional hardware cost in order to distinguish between the live and spoof one. Popular face \cite{1} and iris \cite{7a} biometric artifacts are printed photo, 3D face masks, wrapped printed attack, electronic display of an iris or facial photo \cite{5} as shown in Figure\ref{fig:1}. No matter what kind of presentation attack it is but it can easily jeopardized the security of any biometric based recognition system. Hence, it is important to elude such attacks.

\subsection{Problem statement and motivation} In our work, we are mainly focusing on presentation attack detection for finger dorsal images, which consist of unique line shape patterns on the finger knuckle regions \cite{1b1}. To the best of our knowledge, this is the first attempt in which we present a new prospective for finger dorsal presentation attack detection, collected by light field camera and deep learning mechanism. We have examined the complete finger dorsal structure which is coined as \emph{finger knuckle image} \cite{1b1a} in the literature just as a novel case study, majorly because its anti-spoofing capabilities are not studied at all. However, we have experimentally observed, that the proposed technique is equally applicable to other biometric traits like iris, face or fingerprints. Although, smart-phones and tablets are commonly used in today's technocrat world and they can be easily authenticated biometrically using finger dorsal images in the same manner as the fingerprints are used. Moreover prolonged use of Capecitabine \cite{drug} a drug used for cancer treatment can also remove the fingerprints permanently  as a side effect. In such situations finger-dorsal images can be seen as a suitable  alternative to fingerprint images. One can easily used finger dorsal images for surveillance applications also because it can be captured easily through distance unlike fingerprints. We have taken finger dorsal images just for the case study purpose although the proposed technique can be employed to the other biometric modalities as well. We have used state-of-the-art deep-learning techniques along with a Lytro\cite{lytro} camera for detecting the presentation attack. Here the use of the Lytro camera is limited for training our network only, in  testing scenario we donot require lytro camera. 
\begin{figure*}[!htp]
	\begin{center}
		\includegraphics[width=0.98\linewidth, height=0.73\linewidth]{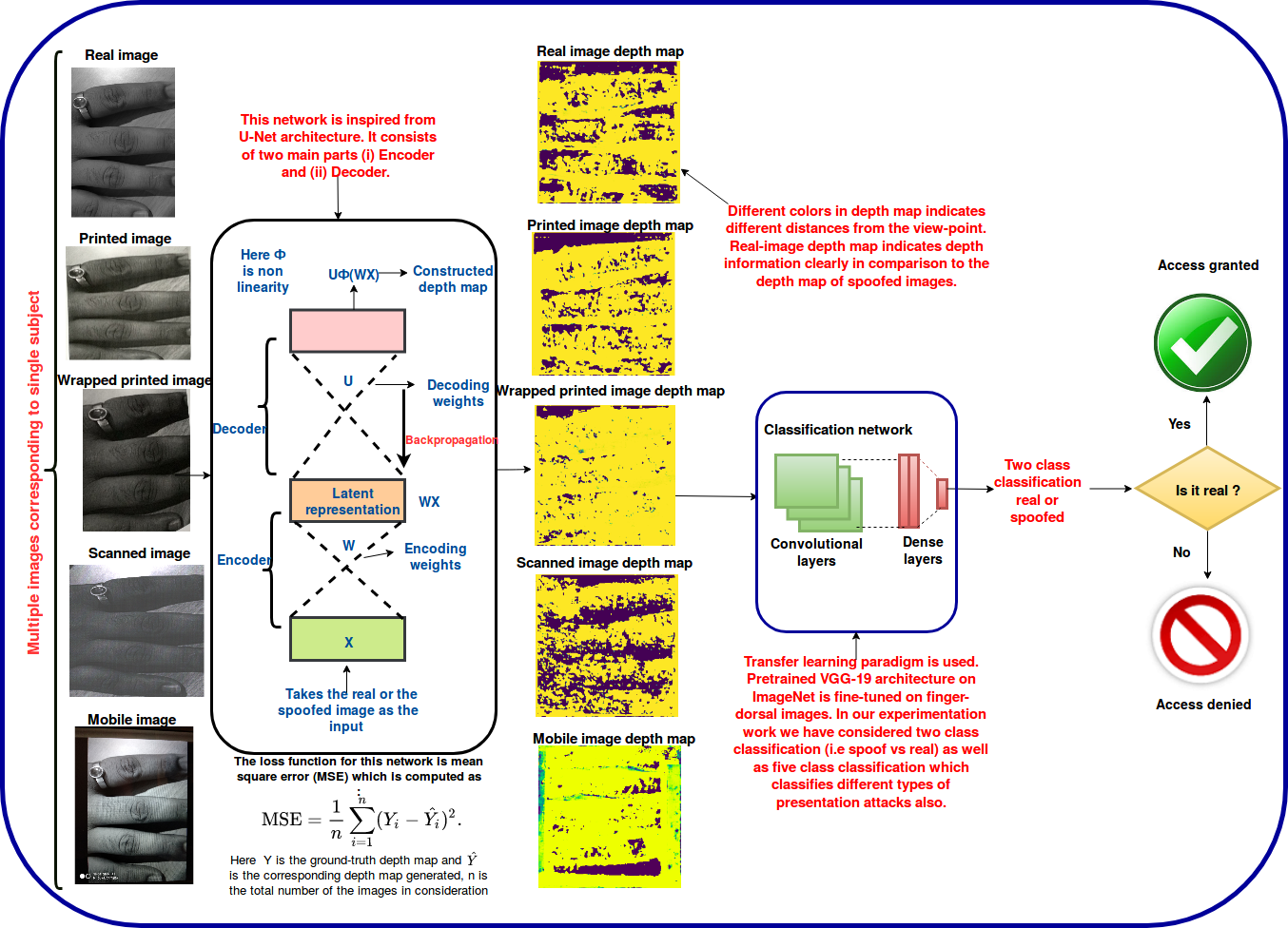}
	\end{center}
	\caption{FDSNet: Finger dorsal spoof detection network overall architecture}
	\label{fig:2}
\end{figure*}

\textbf{FDSNet- Basic architecture detail :} A basic model of our proposed FDSNet (Finger dorsal spoof detection network) is illustrated in Figure \ref{fig:2}. It also depicts the images of presentation attacks that we have considered for finger dorsal images. The two major components of this network are depth extraction network and classification network. More details about these networks are given in the subsequent sections in this paper. It is clearly evident from  Figure \ref{fig:2} that the depth map generated  corresponding to different kinds of attacks differs quite a lot which depicts great learn-ability of our proposed depth extraction network and which is necessary for true classification. Here different colors in depth-map indicates the distance of the surface of the image from different viewpoint. Larger the variation in the dark and the light colors in the depth-map more depth information it contains (as the depth-map of real-image is more informative in Figure \ref{fig:2} as compared to the depth-map of other spoofed images).

\textbf{Presentation attacks considered in this paper :} The major presentation attacks that we discussed in this paper are as follows:
\begin{itemize}
\item Printed image attack: In this kind of attack, images taken from Lytro-camera are printed on A4 size paper using Canon imageClass MF229DW printer. 
\item Wrapped printed image attack: A printed image is wrapped around object having somewhat curvature similar to human hand. This results in different depth information in comparison to a normal printed image. But, it is more challenging to detect this kind of spoofing attack due to depth presence. 
\item Scanned image attack: A printed image is scanned using a Canon imageClass MF229DW machine to create this attack. It means input to scanner are printouts of these images.
\item Mobile image attack: A 2D genuine finger dorsal image is rendered and displayed using Mi-A1 telephoto-lens. 
\end{itemize} 

\subsection{Light field camera and its usefulness}
Depth as a feature is used for differentiating between a real and a fake finger dorsal image. For extracting the depth feature from the finger dorsal  images we have used Lytro-camera\cite{lytro}. It is the first consumer light field camera. Lytro Inc. software is used to import the light field from the camera into the computer in the form of Light Field Picture (LFP) file. This LFP file is a 5D array and its size is $9*9*380*380*4$ where $9$ is the angular resolution, $380$ is the spatial resolution while $4$ is the RGB color channel and the confidence associated with each pixel value. It is very difficult to directly analyze or visualize this 5D file. Open source Light Field Toolbox \cite{toolbox} for MATLAB is used for projecting this 5D light field onto the 2D image. In a conventional camera, the image is formed by summing all the light rays received by each pixel whereas the light field camera can capture the direction of the incident light rays which enables it for surface depth reconstruction as well as for image refocusing. A light-field camera consist of a microlens array, which is embedded in front of the photo sensor that enables it to save the directions of the incident rays. 

\textbf{Usefulness for biometric traits:} Most of the existing studies in the field of biometrics using Lytro camera has been done on detecting spoofed face or fingerprint images mainly because of the large depth variation \cite{2}. However, finger dorsal image which comprises of convex shaped line patterns at knuckle regions \cite{1b1a} and don't have much variation in the depth. As a result, it becomes a quite challenging problem to detect real one from a fake finger dorsal image based on depth features. As far as the presented work is concerned, we  have analyzed raw light field photographs for detecting spoofed finger knuckle images. Here, each photograph is composed of combinations of several images called microlens images. Different light distributions are represented by microlens images depending upon the location of the focal plane and thus it helps in estimating the presence of depth in an image. The real difference between an original finger dorsal and fake finger dorsal image can be depicted easily by the depth information. 

\subsection{Major contributions}
The major contribution of this paper is three fold  which are summarized as follows:
\begin{itemize}
\item  An indigenous in-house dataset for finger dorsal images is being collected using Lytro-camera. This dataset consists of $196$ real images and $784$ spoof images corresponding to $33$ subjects.
\item Transfer learning paradigm is used to train a U-Net based autoencoder architecture for generating depth-map corresponding to input finger-dorsal image.
\item A deep convolutional neural network architecture is used that is capable of differentiating between the depth map of a real and  fake finger-dorsal image. 
\end{itemize}

\section{Related work and important findings}
\begin{small}
 \begin{table*}[!ht]
 \centering
 
 \begin{center}\begin{tabular}{|p {2.2cm}|p {5cm}|p {2.8cm}|p {3.5cm}|p {1.4cm}|}
\hline
\textbf{Methods} &\textbf{Novelty}  &\textbf{Techniques used} & \textbf{Type of spoof} & \textbf{Sensor}\\
 \hline
Raghavendra et al. \cite{3}& Introduced a new face artifact database and extracted depth information& LBP, Hamming distance & Photo print, electronic screen & Lytro\\
 \hline
Moghaddam et al. \cite{9}& Presented a review and benchmarking study on light field based PAD solutions & Histogram of Disparity Gradients & Presentation attack & NA\\
 \hline
Moghaddam et al. \cite{13}& Introduced a Lenslet Light Field Face Spoofing
Database & Light field angular LBP, SVM & Printed, wrapped, laptop, tablet and mobile & Lytro ILLUM\\
 \hline
Kim et al. \cite{4}& Presented a liveness detection approach based on extraction of edge and ray difference features & Sobel operator, PCA, LBP, SVM & Print and tablet & Lytro
 \\
 \hline
\end{tabular}
\caption{Comparative study of existing light field camera based spoof detection}
\end{center}
\label{tab:1}
\end{table*}
\end{small}

Light field camera originates from the concept of plenoptical function and has come into the spotlight, especially with the emergence of camera technology and ongoing research in the field of image processing \cite{1a}. In early 16th century, the idea of light-field imaging was proposed when its strength for acquisition of a photograph was evaluated \cite {1a}. Thereafter, a continuous growth in this field has been observed in terms of sensor technology as well as its usages in different applications \cite{1b}. For example, the use of Kinect camera to extract image information, and near-infrared acquisition system to recognize the image under various illumination at a distance were studied quite extensively for biometrics applications \cite{3}. In addition, thermal and near infrared sensors are some representative examples to develop anti-spoofing capabilities \cite{1}. They can solve the vulnerability, but they do not have merits in terms of cost and commercialization \cite{4}. In order to overcome the limitations of such conventional cameras, light field camera has been applied in various research fields. For example, Lytro, and Raytrix are the most commercialized variants of this technology \cite{4}. However, light field cameras demand serious computing power, expensive price, and larger size which limit its usability in biometric applications up to certain extent. In spite of these challenges, a light field sensor can provide useful information in terms of multiple depth images in a single capture, holding additional information that is quite useful for biometric applications. In \cite {1}, the various characteristics of the camera devices, like light field camera were discussed which used to capture the face image for hardware based PAD approaches. Likewise in \cite{1c}, authors discussed the vulnerability in palm print recognition systems and presented a PAD approach to discriminate the real and fake biometric samples. In \cite {4}, a light field camera based approach for defending spoofing face attacks, like printed 2D facial images was demonstrated, which achieved 94.78\% accuracy over different spoofing attacks. In \cite{2}, authors worked on a problem of multiple face detection at varying distances for which they prepared a multi focus face dataset using light field camera. In \cite{7}, an efficient auto-refocusing iris imaging solution for light field cameras is proposed that selected the best focused image from multiple focus images. In \cite{5}, authors investigated the additional feature information that could be extracted from depth images like face and iris using super resolution mechanism, collected by light field camera. In \cite{8}, authors put efforts to develop light field camera (Lytro ILLUM lenslet) based ear database i.e., Lenslet, consisted of 536 depth images collected from 67 subjects. In \cite{9}, authors presented a review and benchmarking analysis in the literature on light field based face presentation attack detection solutions. In \cite{10}, a 2D fake pedestrian recognition method was proposed for which a pedestrian dataset using light field camera was constructed, consisted of 1000 samples. In \cite{13s}, authors presented a multi-task CNN approach that performed iris detection and iris presentation attack detection efficiently. A few important state-of-art studies are depicted in Table 1.

\section{Proposed FDSNet: Methodology} In the proposed work we have taken finger-dorsal images just for the case study purpose although it is equally applicable to other biometrics  having depth information like face or ear. The obvious differentiating factor between a real and a fake finger-dorsal image is the depth feature. In order to use depth as a distinguishing factor we have collected our own dataset of finger-dorsal images using Lytro-camera. The expensive cost of the light field camera makes it difficult to use it in many applications. Here in our case, also we have used it only for capturing the depth information which is necessary while training the network and not required during testing. Our proposed architecture consists of two subparts (i) depth extraction network (ii) deep classification network. 

\subsection{Depth feature extraction network}
\begin{figure*}[!htp]
	\begin{center}
		\includegraphics[width=0.94\linewidth,height=0.29\linewidth]{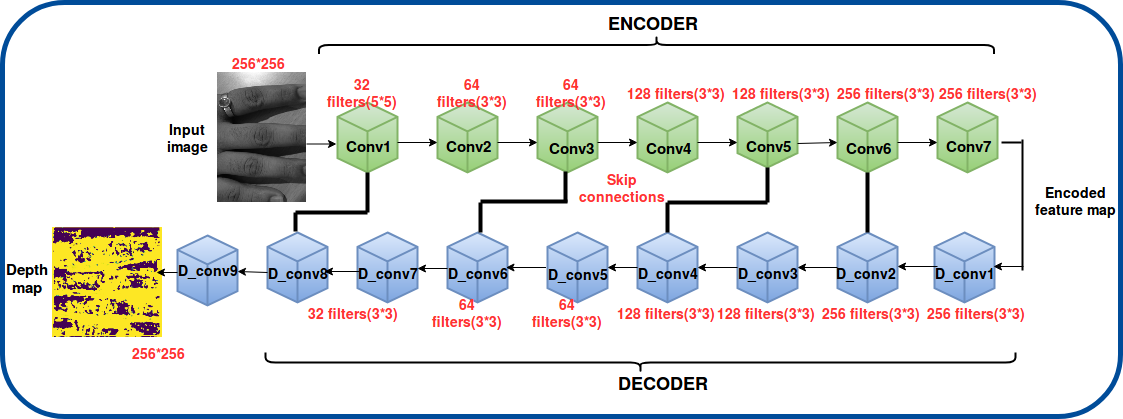}
	\end{center}
	\caption{Depth extraction network architecture}
	\label{fig:3}
\end{figure*}

For extracting the depth feature of our finger dorsal images we have used an architecture inspired from U-Net\cite{unet} architecture. It is a kind of autoencoder with the presence of skip connections in it. Autoencoders are widely used in the literature for many applications like data compression, image-denoising and so on. One of the major advantage of autoencoders is that they can be used for pre-training the deep neural networks which are otherwise difficult to train. The classic U-Net architecture is an end-to-end encoder decoder network that has been used for biomedical image segmentation and consists of $23$ convolutional layers.
Here, our task is not image segmentation but the generation of depth-map. In our case input image to the encoder is a finger dorsal image (it can be either of the two the real one or the generated spoof) and the decoder needs to generate the corresponding depth-map of it. The proposed architecture is depicted by Figure\ref{fig:3}. Here the first part is the encoder part which encodes the input image into features extracted at multiple levels by applying operation like convolution. The second part of the depth extraction network is the decoder part which consists of upsampling and concatenation followed by regular convolution operation. The spatial information in the encoder has been passed on to the decoder part along with the incoming up-sampled features that enables it to generate more location precise depth information.
\subsubsection{Network design and architecture}
The proposed depth extraction network has been trained specifically for depth extraction. It mainly consists of two parts Encoder(E) and Decoder(D) which are described in more detail in the following section.

\textbf{Encoder(E)} The objective function of the encoder network (E) is to map the input finger dorsal image say X to a latent representation say L as:
\begin{equation}
L = \phi(WX)
\end{equation}
Here, W are the encoding weights and $\phi$ is the non-linearity. It is the case when only one hidden layer is used. In reality like in our case as shown in Figure\ref{fig:3} we have used stacked encoder consisting of many hidden units that enables it to learn finer granular level details. In such cases latent representation is written as:
\begin{equation}
L = \phi(W_{l}\circledast(---\phi(W_{2}\circledast(\phi(W_{1}\circledast X)))))
\end{equation}
Here, we have considered \textit{l} hidden units and $W_{l}$ is the encoding weight corresponding to $l^{th}$ unit, $\circledast$ stands for convolution operator. During the designing of the encoder part we have used $7$ convolution layers with ReLU as an activation function. One of the reason of using ReLU is that it avoids and rectifies vanishing gradient problem. In the first convolutional layer we have used $5*5$ filter in order to aggregate global information later on we want more precise depth information locally hence, we have used $3*3$ filter in all subsequent convolutional layers. More details about the number of filters used in each layer is shown in Table 2.

\begin{table}[!htp]
\centering

\label{table2}
\begin{tabular}{|p{0.9cm}|p{0.9cm}|p{0.9cm}|p{1.1cm}|p{0.9cm}|p{0.9cm}|}
\hline
\multicolumn{3}{|c|}{\textbf{Encoder}} & \multicolumn{3}{c|}{\textbf{Decoder}} \\ \hline
\textbf{Layer} & \textbf{Kernel size} & \textbf{No of filters} & \textbf{Layer} & \textbf{Kernel size} & \textbf{No of filters} \\ \hline
Conv1 & 5*5 & 32 & D\_conv1 & 3*3 & 256 \\ \hline
Conv2 & 3*3 & 64 & D\_conv2 & 3*3 & 256 \\ \hline
Conv3 & 3*3 & 64 & Cat1 & - & - \\ \hline
Conv4 & 3*3 & 128 & D\_conv3 & 3*3 & 128 \\ \hline
Conv5 & 3*3 & 128 & D\_conv4 & 3*3 & 128 \\ \hline
Conv6 & 3*3 & 256 & Cat2 & - & - \\ \hline
Conv7 & 3*3 & 256 & D\_conv5 & 3*3 & 64 \\ \hline
 -&-  &-  & D\_conv6 & 3*3 & 64 \\ \hline
 -&-  &-  & Cat3 & - & - \\ \hline
 -&  -&  -& D\_conv7 & 3*3 & 32 \\ \hline
 -& - & - & D\_conv8 & 3*3 & 32 \\ \hline
- & - & - & Cat4 & - & - \\ \hline
 - & - & - & D\_conv9 & 3*3 & 1 \\ \hline
\end{tabular}
\caption{Depth extraction network details}
\end{table}

\textbf{Decoder(D)} Here, the objective function of the decoder network (D) is to map the latent-representation to the corresponding depth-map which can be represented as:
\begin{equation}
X_{depth\_map}= U\phi(WX)
\end{equation}

Here, U = ${W}^\top$ these are the decoding weights. In general decoder with multiple hidden units is mathematically abbreviated as:
\begin{equation}
X_{depth\_map} = \phi(U_{l}\circledast(---\phi(U_{2}\circledast(\phi(U_{1}\circledast X)))))
\end{equation}

Here, we have considered \textit{l} hidden units and $U_{l}$ is the decoding weight corresponding to $l^{th}$ unit. In traditional regeneration work often symmetric-encoder decoder structure is used but here our task is not to regenerate the same input but to generate the corresponding depth-map. Hence, experimentally we have found that asymmetric structure works best in our case.

\subsubsection{Network training} In our collected dataset we have only $196$ real images and training a U-Net based architecture on such small dataset is not possible. In order to overcome this problem we have taken the advantage of transfer-learning paradigm. We have first trained our depth extraction network on \cite{house} dataset consisting of $2284$ images and then fine-tuned it on our dataset using only $196$ images. For training our depth extraction network we have used the depth-map generated corresponding to the images that we have collected using Lytro-camera. Here we have used Lytro-camera because it captures depth-information more accurately than the traditional camera. We have used it only for training purpose, for testing purpose we have not used it because it is too expensive to be deployed in real-world applications.

\textbf{Loss function :} The loss function for this network is mean square error which is written as :
\begin{equation}
MSE = 1/n \sum_{i=1}^{n}(Y_i -\hat{Y_{i}})^2
\end{equation}
Here, $Y_i$ is the ground truth depth-map of image i and $\hat{Y_i}$ is the predicted depth-map of image i and n is the total number of images under consideration. Although this loss is sensitive to outliers, but it gives a more stable and closed form solution (by setting its derivative to $0$).

\textbf{Network justification :} A U-Net based architecture helps in retaining high level semantic information while maintaining low-level details as well. The basic intuition behind its architecture is that copying low-level intrinsic features to the corresponding high-level features actually facilitates backward propagation as well as it also creates a path for information dissemination that allows signals to propagate between low and high levels in a better and easy manner and thus helps in generating more accurate depth-maps. Moreover, the skip connections present in our network helps in back propagating the gradients to bottom layers and disseminate image information to the top layers. These two prominent characteristics helps in end-to-end mapping of input image to the corresponding depth-map. One can easily visualize this by the  depth-maps generated by different types of attacks in Figure \ref{fig:2}.
\subsection{CNN based classification network} Once the depth-map is generated the next task is to differentiate between the depth-map of a real-finger dorsal image and the fake one. For this task we have taken the advantage of CNN based classifiers. In our task we have used state-of-the-art  VGG-$19$ \cite{vgg} architecture trained on ImageNet\cite{imagenet} dataset. Here, we have taken the benefit of transfer learning paradigm and fine-tuned VGG network for our dataset . We basically try to exploit what has been learned in one task to improve generalization in another. One of the main reason of using this paradigm is that we donot have much data to train our own CNN based network from scratch and this technique helps us to leverage the labelled data of the task it was initially trained on. Another reason of using this architecture is that it is one of the most validated network on various benchmark datasets. Traditional VGG-19 architecture have in total $22$ layers out of which there are $16$ Conv layers, $5$ pooling layers with no associated weight and $1$ input layer. While tweaking this architecture for our task we have freezed all the Conv layers except the last Conv layer and added $2$ dense layers. We have experimentally decided  the best architecture for our problem. Moreover, we have also added two dense layers in-order to avoid any kind of over-fitting in our network.

\textbf{Network implementation details:} Both the proposed depth-extraction network as well as classification network has been implemented using python and Keras \cite{keras} library using tensorflow \cite{tensorflow} as the back-end. All the implementations has been done on Intel(R) Xenon(R) CPU E5-2630 with 32 GB RAM and NVIDIA Tesla K40C GPU card with on card RAM is 12GB.

\section{Performance analysis}
This section presents the description of in-house database, testing strategy, and performance matrices for the evaluation of finger dorsal spoofing detection system.

\subsection{Data acquisition set-up and image description}
To best of our knowledge, there is no finger dorsal database available in public domain which is collected by a light field camera. In this study, we made efforts to capture light field photographs of real finger dorsals with a Lytro camera and named this dataset as  \textbf{Lytro Finger Dorsal} (LFD) dataset. All images were taken in the indoor lighting condition. The background was beige colored with wooden texture. The distance between the camera-lens and the subject hand was kept around 60 cm. We have collected  a total of $196$ real images from $33$ subjects in our experimentation work. In addition to develop spoof finger dorsal images, we have created $196$ samples corresponding to different presentation attacks like mobile, printed, wrapped, and scan. Thus in total, we get $980$ images out of which $196$ are real and remaining as spoof.
\subsection{Performance matrices and testing protocol}
To evaluate the performance of our proposed method on in-house LFD database, the standard matching criteria followed by most of the existing works in literature \cite{13} is used. For experimentation, database is mainly categorized into two groups: training and testing sets. Each set comprises of 50\% of the total images. In the proposed finger dorsal spoof detection following measures are used: 
\begin{itemize}
\item \textbf{False acceptance rate (FAR)}: the proportion of spoof images which are classified as real
\begin{equation}
FAR = \frac {FP} {FP +TN}
\end{equation}
where FP and TN stands for false positive and true negative respectively.
\item \textbf{False rejection rate (FRR)}: the proportion of real images which are classified as spoof.
\begin{equation}
FRR = \frac{FN}{TP+FN}
\end{equation}
where FN and TP stands for false negative and true positive respectively.
\item \textbf{Total error rate (TER)}: summation of FAR and FRR.
\begin{equation}
TER = FAR+FRR
\end{equation}
\item \textbf{Half total error rate (HTER)}: half of TER.
\begin{equation}
HTER = \frac {TER} {2}
\end{equation}
\item \textbf{Correct recognition rate (CRR)}: the ratio of correctly classified test images upon the total number of test images.

\begin{equation}
CRR = \frac{TP+TN}{TP+TN+FP+FN}
\end{equation}
\end{itemize}

\subsection{Experiments and discussions }
\begin{figure}[!htp]
	\begin{center}
		\includegraphics[width=0.98\linewidth, height=0.63\linewidth]{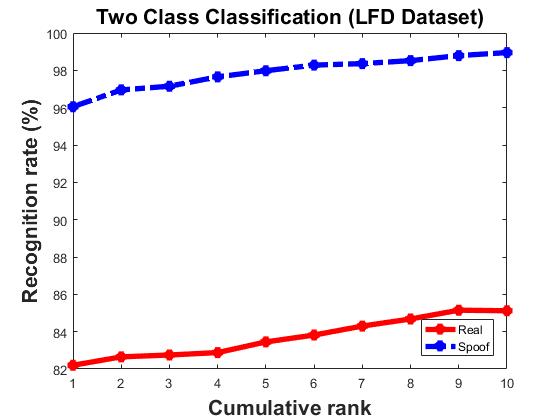}
	\end{center}
	\caption{Two Class Classification performance}
	\label{fig:ch1a}
\end{figure}

We examine the classification performance in accordance with the generated depth maps and types of spoofing attacks. For this, we have fine tuned the pre-trained VGG network on our finger dorsal dataset consisting of $980$ images, with different hyper parameter settings. In order to avoid the over-fitting in the network, data augmentation has been performed during training. The two type of experiments are given below: 

\textbf{Two class Classification :} In first experiment, the task is to classify the finger dorsal image into discrete outcomes or classify the image into either a group of real or fake finger dorsal image. It is also named as binary classification or one-vs-one (ovo). The dataset was divided into independent training and testing sets, as discussed above. This approach gets to a CRR for real and fake images as 82.65\% and 96.93\% respectively. In order to perform fine-tuning, we have freezed  $16$ Conv  layers of the VGG19 model, and trained only the last Conv layer along with two dense layers after 50 epochs. Learning rate was kept $0.0001$ and SGD optimizer was used. The results with these hyper-parameter setting are presented in Table \ref{tab:b1} which clarifies that our approach is performing quite well for detecting spoof prints while the performance on real images is not obtained so good. Although, we can easily improve the results by tweaking the hyper-parameters of CNN n/w but our main focus is to get high performance for spoof detection. The CMC plot for two class classification is shown in Figure \ref{fig:ch1a}. 
\begin{table}[]
\centering

\begin{tabular}{|c|c|c|c|}
\hline
\textbf{Mode}            & \textbf{TER} & \textbf{HTER} & \textbf{CRR} \\ \hline
\multicolumn{4}{|c|}{\textbf{Two class classification}}\\
\hline
\textbf{Real} & NA   & NA  & 82.65    \\ \hline
\textbf{Spoof} & 0.777    & 0.3888   & 96.93   \\ \hline
\multicolumn{4}{|c|}{\textbf{Multi-class classification}}\\
\hline
\textbf{Real}     & NA  & NA & 96.86      \\ \hline
\textbf{Mobile attack}     & 0  & 0 & 98.97      \\ \hline
\textbf{Print attack}   & 0.061& 0.030 & 40.81\\ \hline
\textbf{Wrapped print attack}    & 0.030 & 0.015 & 74.48\\ \hline
\textbf{Scan attack}     & 0.142& 0.071 & 84.69      \\ \hline
\end{tabular}
\caption{Performance analysis}
\label{tab:b1}
\end{table}

\begin{figure}[!htp]
	\begin{center}
		\includegraphics[width=0.98\linewidth, height=0.63\linewidth]{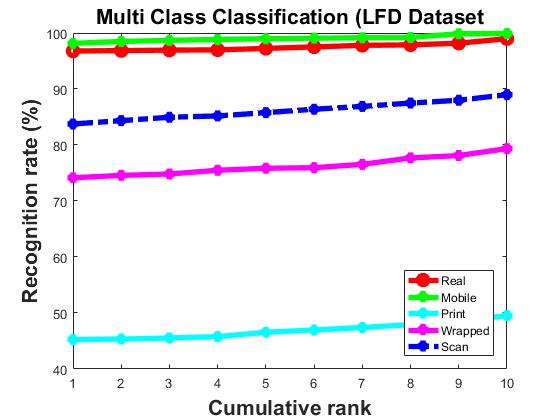}
	\end{center}
	\caption{Multi Class Classification performance}
	\label{fig:ch1a1}
\end{figure}

\textbf{Multi-class Classification :} The task of classifying instances into one of three or more classes is called as multi class classification. It is also named as one-vs-all (ova). In this experiment, unlike two class classification, our aim is not only to classify into real or spoof images but also to identify the type of spoof-attack. All the results for multi-class classification are listed in Table \ref{tab:b1}. From this, it can be observed that the obtained results for print attacks and wrapped print attack are not so good. This is mainly because the depth map generated for these two classes on most of the images are quite similar. Due to this, they fall with in each other class frequently and thus results in poor performance. 
On the other side we have achieved outrageous results on mobile attack due to the dissimilarity in the depth-feature information. This approach gets to a CRR of 98.97\% after 200 epochs. With this it justifies the importance  of this work as in this digital era  spoof attacks are of prime concern for biometric based authentication systems. We have also tested our network on real images captured through normal camera and got similar results. Thus, this experimentation eliminates the need of lytro camera during testing for real world deployment. We only need it for training our network. The CMC plot for multi class classification is shown in Figure \ref{fig:ch1a1}.

\section{Conclusion and future directions} 

In this paper, a CNN based spoof detection system for finger dorsal images is suggested, just for the case study purpose. In particular, we have extracted the depth feature maps of light field finger dorsal images and examined the classification performance for four different type of presentation attacks. In future, we would make efforts to enhance the same in-house dataset for more individuals so as to train the proposed CNN classification model from scratch which will be exclusively trained particularly for detecting spoof from real ones. Additionally, we would like to test our approach on other biometric traits like face and ear, which have self tendency of depth information. Also, we would consider the same problem on more challenging video based print attacks.

{\small
\bibliographystyle{ieee}
\bibliography{gaurav}
}

\end{document}